\documentclass[10pt,twocolumn,letterpaper]{article}

\usepackage{cvpr}
\usepackage{times}
\usepackage{epsfig}
\usepackage{graphicx}
\usepackage{caption}
\usepackage[font=small]{caption}
\usepackage{amsmath}
\usepackage{amssymb}

\usepackage[pagebackref=true,breaklinks=true,letterpaper=true,colorlinks,bookmarks=false]{hyperref}
\DeclareMathAlphabet{\pazocal}{OMS}{zplm}{m}{n}

\usepackage{enumitem}
\usepackage{cite}   
\usepackage{enumitem}
\usepackage{graphicx}
\usepackage{subfloat}
\usepackage{tabularx}
\usepackage{adjustbox}
\usepackage{amssymb}
\usepackage{multirow}
\usepackage[flushleft]{threeparttable}
\usepackage{subfigure}
\usepackage{tablefootnote}
\usepackage{lipsum}
\newcommand\blfootnote[1]{%
  \begingroup
  \renewcommand\thefootnote{}\footnote{#1}%
  \addtocounter{footnote}{-1}%
  \endgroup
}

\newcommand{\figureref}[1]{Figure~\ref{#1}}
\newcommand{\tabref}[1]{Table.~\ref{#1}}
\newcommand{\tabfref}[1]{Table~\ref{#1}}

\renewcommand{\paragraph}[1]{\vspace{1mm}\noindent\textbf{#1}}

\renewcommand{\ie}{\textit{i.e.}}
\renewcommand{\eg}{\textit{e.g.}}

\cvprfinalcopy 

\def\cvprPaperID{5785} 

\ifcvprfinal\fi
\begin{document}
\twocolumn[{%
\renewcommand\twocolumn[1][]{#1}%

\title{Video Panoptic Segmentation}

\author{Dahun Kim\textsuperscript{1,$\dagger$} \quad 
Sanghyun Woo\textsuperscript{1,$\dagger$} \quad
Joon-Young Lee\textsuperscript{2} \quad
In So Kweon\textsuperscript{1}\\ \\
\textsuperscript{1}KAIST \qquad \textsuperscript{2}Adobe Research
}
\maketitle

\begin{center}
    \centering
    \includegraphics[width=\textwidth,height=5cm]{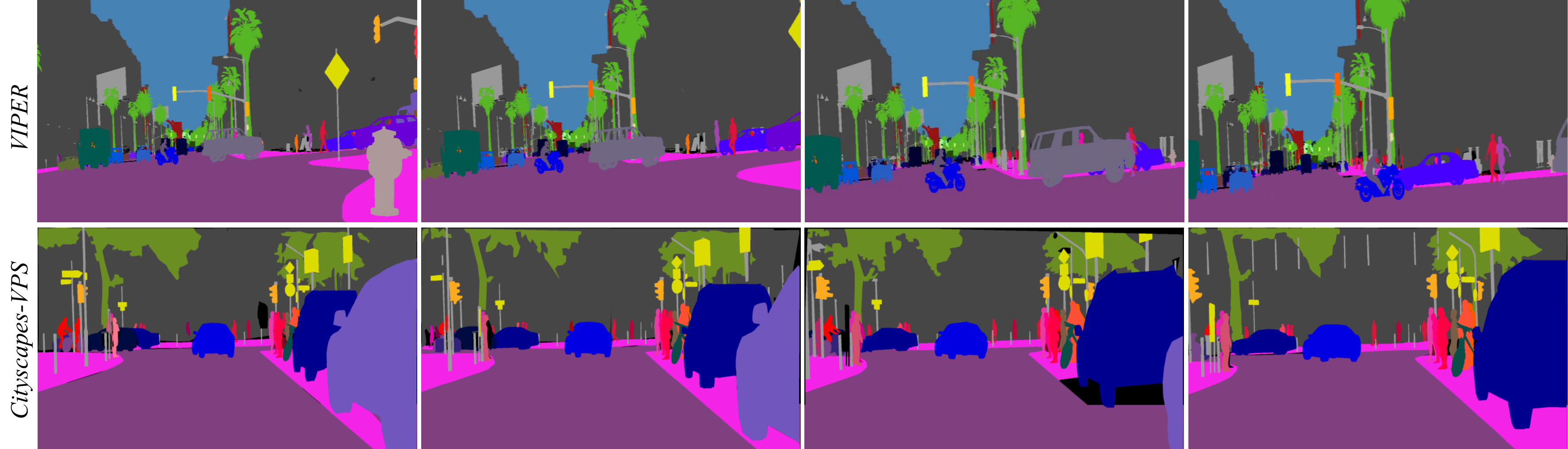}
    \vspace{-6mm}
    \captionof{figure}{Example video sequences of reformatted VIPER and newly created Cityscapes-VPS annotations for video panoptic segmentation. }
    \label{fig:teaser}
\end{center}%
}]
%
\begin{abstract}
    \vspace{-1mm}
     Panoptic segmentation has become a new standard of visual recognition task by unifying previous semantic segmentation and instance segmentation tasks in concert. In this paper, we propose and explore a new video extension of this task, called video panoptic segmentation. The task requires generating consistent panoptic segmentation as well as an association of instance ids across video frames. To invigorate research on this new task, we present two types of video panoptic datasets. The first is a re-organization of the synthetic VIPER dataset into the video panoptic format to exploit its large-scale pixel annotations. The second is a temporal extension on the Cityscapes val. set, by providing new video panoptic annotations (Cityscapes-VPS). Moreover, we propose a novel video panoptic segmentation network (VPSNet) which jointly predicts object classes, bounding boxes, masks, instance id tracking, and semantic segmentation in video frames. To provide appropriate metrics for this task, we propose a video panoptic quality (VPQ) metric and evaluate our method and several other baselines. Experimental results demonstrate the effectiveness of the presented two datasets. We achieve state-of-the-art results in image PQ on Cityscapes and also in VPQ on Cityscapes-VPS and VIPER datasets. The datasets and code are available at \url{https://github.com/mcahny/vps}. \blfootnote{\textsuperscript{$\dagger$} This work was done during an internship at Adobe Research.}
\end{abstract}

\section{Introduction}

As an effort to unify existing recognition tasks, object classification, detection, and segmentation and to leverage the possible complementariness of these tasks into a single complete task, Kirillov~\etal~\cite{kirillov2019panoptic} proposed a holistic segmentation of all foreground instances and background regions in a scene and named the task \textit{panoptic segmentation}. Since then, a large number of works~\cite{li2018weakly,kirillov2019panoptic2,li2019attention,xiong2019upsnet,liu2019end,fu2019imp,lazarow2019learning,li2018learning,porzi2019seamless,de2018panoptic,de2019single,yang2019deeperlab} have proposed learning-based approaches to this new benchmark task, confirming its importance to the field.

In this paper, we extend the panoptic segmentation in the image domain to the video domain. Different from image panoptic segmentation, the new problem aims at a simultaneous prediction of object classes, bounding boxes, masks, instance id associations, and semantic segmentation, while assigning unique answers to each pixel in a video. \figureref{fig:teaser} illustrates sample video sequences of ground truth annotations for this problem. Naturally, we name the new task \textit{video panoptic segmentation} (VPS). The new task opens up possibilities for applications that require a holistic and global view of video segmentation such as autonomous driving, augmented reality, and video editing. In particular, temporally dense panoptic segmentation of a video can work as intermediate-level representations for even higher-level video understanding tasks such as temporal reasoning or action-actor recognition which anticipates the behaviors of objects and humans. To best of our knowledge, this is the first work to address video panoptic segmentation problem.

Thanks to the existence of panoptic segmentation benchmarks such as COCO~\cite{lin2014microsoft}, Cityscapes~\cite{Cordts2016Cityscapes}, and Mapillary~\cite{neuhold2017mapillary}, the panoptic \textit{image} segmentation has successfully driven active participation of the community. However, the direction towards the video domain has not yet been explored, probably due to the lack of appropriate datasets and evaluation metrics. While video object/instance segmentation datasets are available these days, no dataset permits direct training of video \textit{panoptic} segmentation (VPS). This is not surprising when considering its extremely high cost of collecting such data. To improve the situation, we make an important first step in the direction of panoptic \textit{video} segmentation, by presenting two types of datasets. First, we adapt the synthetic VIPER~\cite{richter2017playing} dataset into the video panoptic format and create corresponding metadata. Second, we collect a new video panoptic segmentation dataset, named \textit{Cityscapes-VPS}, that extends the public Cityscapes to a video level by providing every five video frames with pixel-level panoptic labels that are temporally associated with respect to the public image-level annotations.

In addition, we propose a video panoptic segmentation network (VPSNet) to provide a baseline method for this new task. On top of UPSNet\cite{xiong2019upsnet}, which is a state-of-the-art method for image panoptic segmentation, we design our VPSNet to take an additional frame as the reference to correlate time information at two levels: pixel-level fusion and object-level tracking. To pick up the complementary feature points in the reference frame, we propose a flow-based feature map alignment module along with an asymmetric attention block that computes similarities between the target and reference features to fuse them into \textit{one-frame} shape. Moreover, to associate object instances across time, we add an object track head~\cite{yang2019video} which learns the correspondence between the instances in the target and reference frames based on their RoI feature similarity. It establishes a baseline for the VPS task and gives us insights into the main algorithmic challenges it presents.

We adapt the standard image panoptic quality (PQ) measure to fit the video panoptic quality (VPQ) format. Specifically, the metric is obtained from a span of several frames, where the sequence of each panoptic segment within the span is considered a single 3D tube prediction to produce an IoU with the ground truth tube. The longer the time-span, the more challenging it is to obtain IoU over a threshold and to be counted as a true-positive for the final VPQ score. We evaluate our proposed method with several other naive baselines using the VPQ metric. 

Experimental results demonstrate the effectiveness of the two presented datasets. Our VPSNet achieves state-of-the-art image PQ on Cityscapes and VIPER. More importantly, in terms of VPQ, it outperforms the strong baseline~\cite{yang2019video} and other simple candidate methods, while still presenting algorithmic challenges of the VPS task.

We summarize the contribution of this paper as follows.
\begin{enumerate}[topsep=0pt,itemsep=0pt]
\item  To our best knowledge, it is the first time that video panoptic segmentation (VPS) is formally defined and explored.

\item  We present the first VPS datasets by re-formatting the virtual VIPER dataset and creating new video panoptic labels based on the Cityscapes benchmark. Both datasets are complementary in constructing an accurate VPS model.

\item  We propose a novel VPSNet which achieves state-of-the-art image panoptic quality (PQ) on Cityscapes and VIPER, and compare it with several baselines on our new datasets.

\item We propose a video panoptic quality (VPQ) metric to measure the spatial-temporal consistency of predicted and ground truth panoptic segmentation masks. The effectiveness of our proposed datasets and methods is demonstrated by the VPQ evaluation.

\end{enumerate}

\section{Related Work}
\label{sec:related}
\paragraph{Panoptic Segmentation: }
The joint task of thing and stuff segmentation is reinvented by Kirillov~\etal~\cite{kirillov2019panoptic} in the form of combining the semantic segmentation and instance segmentation tasks and is named panoptic segmentation. Since then, much research~\cite{li2018weakly,kirillov2019panoptic2,li2019attention,xiong2019upsnet,liu2019end,fu2019imp,lazarow2019learning,li2018learning,porzi2019seamless,de2018panoptic,de2019single,yang2019deeperlab} has been actively gathered to propose new approaches to this unified task, which is now a \textit{de facto} standard of visual recognition task. A naive baseline introduced in~\cite{kirillov2019panoptic} is to train the two sub-tasks separately then fuse the results by heuristic rules. More advanced approaches to this problem present a unified, end-to-end model. Li~\etal~\cite{li2019attention} propose AUNet which leverages mask level attention to transfer knowledge from the instance head to the semantic head. Li~\etal~\cite{li2018learning} suggest a new objective function to enforce consistency between things and stuff pixels when merging them into a single segmentation result. Liu~\etal~\cite{liu2019end} design a spatial ranking module to address the occlusion between the predicted instances. Xiong~\etal~\cite{xiong2019upsnet} introduce a non-parametric panoptic head to predict instance id and resolve the conflicts between things and stuff segmentation. 

\paragraph{Video Semantic Segmentation: }
As a direct extension of semantic segmentation to videos, all pixels in a video are predicted as different semantic classes. However, the research in this field has not gained much attention and not currently popular compared to its counterpart in the image domain. One possible reason is the lack of available training data with temporally dense annotation, as research progress depends greatly on the existence of datasets. Despite the absence of a dataset for Video Semantic Segmentation (VSS), several approaches have been proposed in the literature~\cite{li2018low, shelhamer2016clockwork, zhu2017deep, nilsson2018semantic, jain2019accel}. Temporal information is utilized via optical flow to improve the accuracy or efficiency of the scene labeling performance. Different from our setting, VSS does not require either discriminating object instances or explicit tracking of the objects across frames. Our new \textit{Cityscapes-VPS} is a super-set of a VSS dataset and thus is able to benefit this independent field as well.

\paragraph{Video Instance Segmentation: } 
Even more recently, Yang~\etal~\cite{yang2019video} proposed a Video Instance Segmentation (VIS) problem to extend image instance segmentation to videos. It combines several existing tasks: video object segmentation~\cite{caelles2017one, chen2018blazingly, cheng2017segflow, perazzi2017learning,tokmakov2017learning, yang2018efficient, wug2018fast, oh2019video} and video object detection~\cite{feichtenhofer2017detect, zhu2017flow, zhu2017deep}, and aims at simultaneous detection, segmentation, and tracking of instances in videos. They propose Mask-Track R-CNN which has a tracking branch added to Mask R-CNN~\cite{he2017mask} to jointly learn these multiple tasks. The object association is trained based on object feature similarity learning, and the learned features are used together with other cues such as spatial correlation and detection confidence to track the objects at inference. The first difference to our setting is that VIS only deals with foreground \textit{thing} objects but not background \textit{stuff} regions. Moreover, the problem permits overlaps between predicted object masks and even multiple predictions for a single instance, while our task requires algorithms to assign a single label to all things and stuff pixels. Last but not least, their dataset contains a small number of objects ($\sim 5$) per frame, whereas we deal with a much larger number of objects ($ > 20$ on average), which makes our task even more challenging.

\section{Problem Definition}

\paragraph{Task Format:} For a video sequence with $T$ frames, we set a temporal window that spans additional $k$ consecutive frames. Given a $k$-span snippet $I^{t:t+k}=\{I^t, I^{t+1},..., I^{t+k}\}$, we define a \textit{tube} prediction as a track of its frame-level segments as $\hat{u}_{(c_i, z_i)} = \{\hat{s}^t,..., \hat{s}^{t+k}\}_{(c_i, z_i)}$, for semantic class $c$ and instance id $z$ of the tube. Note that instance id $z_i$ for a \textit{thing} class can be larger than 0, \eg, \textit{car}-0, \textit{car}-1, ... , whereas it is always $0$ for a \textit{stuff} class, \eg, \textit{sky}. All pixels in the video are grouped by such tuple prediction, and they will result in a set of \textit{stuff} and \textit{things} video tubes that are mutually exclusive to each other. The ground truth tube is defined similarly, with a slight adjustment concerning the annotation frequency as described below. The goal of video panoptic segmentation is to accurately localize all the semantic and instance boundaries throughout a video and assign correct labels to those segmented video tubes.

\begin{figure}[t]
\begin{center}
\begin{tabular}{@{}c@{}}
\includegraphics[width=1.0\linewidth]{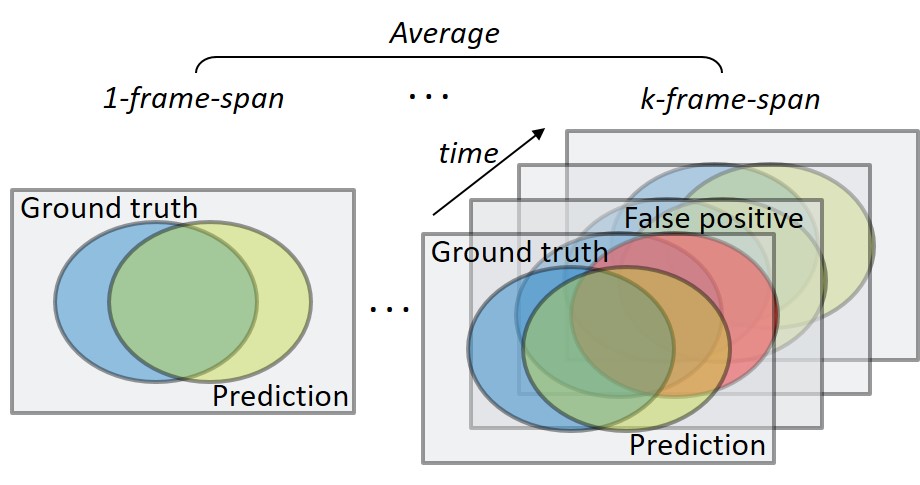} \\
\end{tabular}
\end{center}
\vspace{-6mm}
\caption{\textbf{Tube matching and video panoptic quality (VPQ) metric.} An IoU is obtained by matching predicted and ground truth \textit{tubes}. A frame-level false positive segment penalizes the whole predicted tube to get a low IoU. Each VPQ\textsuperscript{$k$} is computed by sliding the window through a video, and averaged by the number of frames. $k$ indicate the temporal window size. VPQ\textsuperscript{$k$} is then averaged over different $k$ values, to get a final VPQ score. }
\label{fig:vpq_metric}
\vspace{-3mm}
\end{figure}

\paragraph{Evaluation Metric:}
By the construction of the VPS problem, no overlaps are possible among video tubes. Thus, AP metric used in object detection or segmentation cannot be used to evaluate the VPS task. Instead, we borrow the panoptic quality (PQ) metric in image panoptic segmentation with modifications adapted to our new task. 

Given a snippet $I^{t:t+k}$, we denote a \textit{\textbf{set}} of the ground truth and predicted tubes as $\mathcal{U}^{t:t+k}$ and $\hat{\mathcal{U}}^{t:t+k}$. A set of True Positive matches is defined as TP = $\{ (u, \hat{u}) \in \mathcal{U} \times \hat{\mathcal{U}} $ : IoU ($u, \hat{u}) > 0.5$ \}. False Positives (FP) and False Negatives (FN) are defined accordingly. When the annotation is given every $\lambda$ frames, the matching only considers the annotated frame indices $t:t+k:\lambda$ $(start:end:stride)$ in a snippet, \eg, when $k$ = 10 and $\lambda$ = 5, frame $t$, $t$+5 and $t$+10 are considered. We slide the $k$-span window with a stride $\lambda$ throughout a video, starting from frame 0 to the end, \ie, $t$ goes by $0:T-k:\lambda$ (We assume frame 0 is annotated). Each stride constructs a new snippet, where we compute the IoUs, TP, FP and FN as above.

At a dataset level, the snippet-level IoU, $|$TP$|$, $|$FP$|$ and $|$FN$|$ values are collected \textit{across all predicted videos}. Then, the \textit{dataset-level} VPQ metric is computed per each class $c$, and averaged across all classes as,
\begin{equation}
    VPQ^{k} = {1\over N_{classes}} {\sum_c{{{\sum_{(u,\hat{u}) \in TP_{c}}IoU(u,\hat{u})} \over {{|TP_c|} + {1 \over 2} |FP_c| + {1 \over 2} |FN_c|}} }}, \\ 
\end{equation}
where ${1 \over 2} |FP| +  {1 \over 2} |FN|$ in the denominator is to penalize unmatched tubes, as suggested in the image PQ metric.

By definition, $k$ = 0 will make the metric equivalent to the image PQ metric, and $k$ = $T$-1 will construct a set of whole video-long tubes. Any cross-frame inconsistency of semantic or instance label prediction will result in a low tube IoU, and may drop the match out of the TP set, as illustrated in ~\figureref{fig:vpq_metric}. Therefore, the larger window size we have, the more challenging it is to get a high VPQ score. In practice, we include different window sizes $k$ $\in$ $\{0,5,10,15\}$ to provide a more comprehensive evaluation. The final VPQ is computed by averaging over $K$ = 4 as, 
$VPQ = {1\over{K}}     {\sum_{k}{VPQ}^k}$.

Having different $k$ values enables a smooth transition from the existing image PQ evaluation to videos, encouraging the image-to-video transition of further technical developments for this pioneering field to leap forward.

\paragraph{Hyper-parameter:} We set $k$ as a user-defined parameter. Having such a fixed temporal window size regularizes the difficulty of IoU matching across video samples of different lengths. On the other hand, the difficulty of matching whole $T$-long tubes, extremely varies with the video length, \eg, when $T$ = 10 and $T$ = 1000.

We empirically observed that, in our Cityscapes-VPS dataset ($\lambda = 5$), many object associations are disconnected by significant scene changes when $k > 15$. Given a new annotation frequency (1/$\lambda$), the $k$ shall be reset, which will accordingly set a level of difficulty for the dataset.

\begin{figure*}[t]
\begin{center}
\begin{tabular}{@{}c@{}}
\includegraphics[width=0.85
\linewidth]{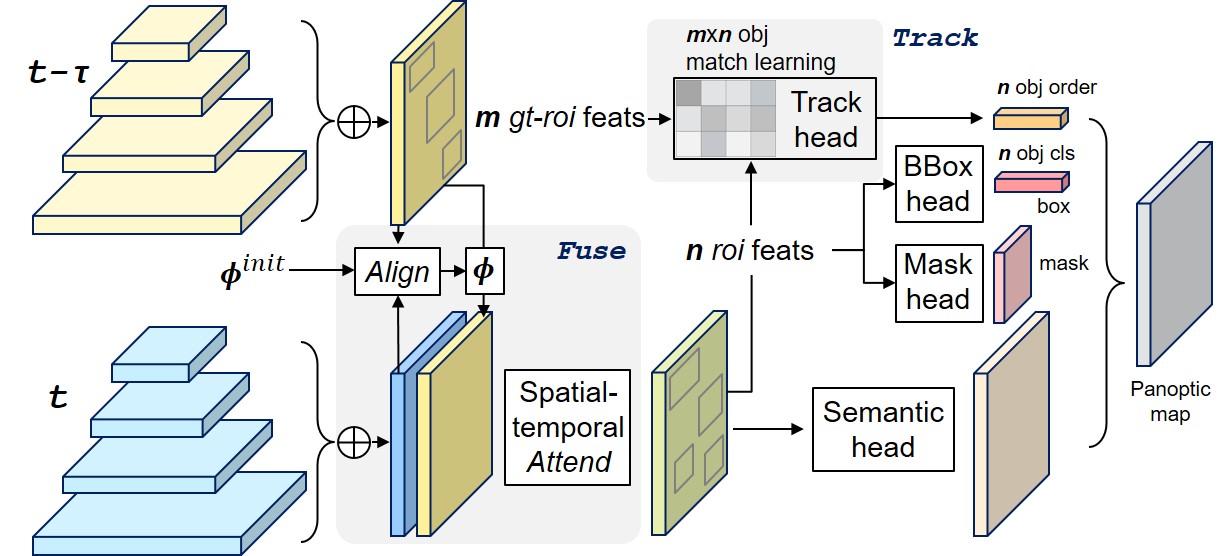} \\
\end{tabular}
\end{center}
\vspace{-7mm}
\caption{\textbf{Overall architecture of our VPSNet.} }
\vspace{-3mm}
\label{fig:vpsnet}
\end{figure*}

\section{Dataset Collection}

\paragraph{Existing Image-level Benchmarks:}
There are several public datasets which have dense panoptic segmentation annotations: Cityscapes~\cite{Cordts2016Cityscapes}, ADE20k~\cite{zhou2019semantic}, Mapillary~\cite{neuhold2017mapillary}, and COCO~\cite{lin2014microsoft}. However, none of these datasets matches the requirement for our video panoptic segmentation task. Thus, we need to prepare a suitable dataset for the development and evaluation of video panoptic segmentation methods. We pursue several directions when collecting VPS datasets. First, both the quality and quantity of the annotation should be high, of which the former is a common problem in some of the existing polygon-based segmentation datasets and the latter is limited by the extreme cost of panoptic annotations. More importantly, it should be easily adaptable to and extensible from the existing image-based panoptic datasets, so that it can promote the research community to seamlessly transfer the knowledge between the image and video domains. With the above directions in mind, we present two VPS datasets by 1) reformatting the VIPER dataset and 2) creating new video panoptic annotations based on the Cityscapes dataset.

\paragraph{Revisiting VIPER dataset:}
To maximize both the quality and quantity of the available annotations for the VPS task, we take advantage of the synthetic VIPER dataset~\cite{richter2017playing} extracted from the GTA-V game engine. It includes pixel-wise annotations of semantic and instance segmentations for 10 \textit{thing} and 13 \textit{stuff} classes on 254K frames of ego-centric driving scenes at $1080\times1920$ resolution. As shown in~\figureref{fig:teaser}-(top row), we tailor their annotations into our VPS format and create metadata in a popular COCO style, so that it can be seamlessly plugged into recent recognition models such as Mask-RCNN~\cite{he2017mask}.

\begin{table}[t]
\centering
\resizebox{0.9\linewidth}{!}{%
\begin{tabular}[b]{l|c|c|c|c}
        \hline 
                            & YT-VIS    & City      & re-VIPER     & City-VPS  \\ \hline\hline
                Videos      & 2540      & 3475		&  124      & 500       \\ \hline
                Frames      & 108k      & 3475		&  184k     & 3000      \\ \hline
                Things      & 40        & 8         &  10       & 8         \\ \hline
                Stuff       &  x        & 11        &  13       & 11        \\ \hline
                Instances   & 4297      & 60 K      &  31 K     & 10 K      \\ \hline
                Masks       & 115 K     & 60 K      &  2.8 M    & 56 K       \\ \hline
                Temporal    &\checkmark & x         &\checkmark &\checkmark \\ \hline
        Dense (Panoptic)    & x         &\checkmark &\checkmark &\checkmark \\ \hline
\end{tabular}
}
\caption{High-level statistics of our reformatted VIPER and new Cityscapes-VPS (additional to the original Cityscapes) with previous video instance / semantic segmentation datasets. YT-VIS and City stands for YouTube-VIS and Cityscapes respectively.}
\vspace{-5mm}
\label{tab:stat}
\end{table}

\paragraph{Cityscapes-VPS:}
Instead of building our dataset from scratch in isolation, we build our benchmark on top of the public Cityscapes dataset~\cite{Cordts2016Cityscapes}, which is the most popular dataset for panoptic segmentation, together with COCO. It consists of image-level annotated frames of ego-centric driving scenarios, where each labeled frame is the 20th frame in a 30 frame video snippet. There are 2965, 500, and 1525 such sampled images paired with dense panoptic annotations for 8 \textit{thing} and 11 \textit{stuff} classes for training, validation, and testing, respectively. 
Specifically, we select the validation set to build our own video-level extended dataset. We select every five frames from each of the 500 videos, \ie, 5, 10, 15, 20, 25, and 30-th frames, where the 20-th frame already has the original Cityscapes panoptic annotations. For the other 5 frames, we ask expert turkers to carefully label each pixels with all 19 classes and instance ids to be consistent with the 20-th frame as reference. It is also asked to have similar level of pixel quality, as shown in~\figureref{fig:teaser}-(bottom row). Our resulting dataset provides additional 2500 frames of dense panoptic labels at $1024\times2048$ resolution that temporally extend the 500 frames of the Cityscapes labels. The new benchmark is referred to as \textit{Cityscapes-VPS}.

Our new dataset \textit{Cityscapes-VPS} is not only the first benchmark for video panoptic segmentation but also a useful benchmark for other vision tasks such as video instance segmentation and video semantic segmentation; the latter has also been suffering lack of well-established video benchmark. We show some high-level statistics of the reformatted VIPER and new Cityscapes-VPS, and related datasets in~\tabref{tab:stat}.

\section{Proposed Method}
Unlike static images, videos have rich temporal and motion context, and a VPS model should faithfully use this information to capture the \textit{panoptic} movement of all \textit{things} and \textit{stuff} classes in a video. We propose a video panoptic segmentation network (VPSNet). Given an input video sequence, VPSNet performs object detection, mask prediction, tracking, and semantic segmentation all simultaneously. This section describes our network architecture and its implementation in detail.

\subsection{Network Design}
\paragraph{Overview:}
By the nature of the VPS task, temporal inconsistency in any of the class label and instance id will result in low video quality of these panoptic segmentation sequences. More strict requirements are therefore in place for the \textit{thing} classes. With this consideration in mind, we design our VPSNet to use video context in two levels: pixel level and object level. The first is to leverage neighboring frame features for the downstream multi-task branches and the second is to explicitly model cross-frame instance association specifically for tracking. Each module for feature fusion and object tracking is not totally new in isolation, but they both are jointly used for the first time for the task of video panoptic segmentation. We call each of them \textit{Fuse} and \textit{Track} module throughout the paper. The overall model architecture is shown in ~\figureref{fig:vpsnet}.

\paragraph{Baseline:}
We build upon an image-level panoptic segmentation network. While not being sensitive to any specific design of a baseline network, we choose the state-of-the-art method, UPSNet~\cite{xiong2019upsnet}, which adopts Mask R-CNN~\cite{he2017mask} and deformable convolutions~\cite{dai2017deformable} for instance and semantic segmentation branches respectively with a panoptic head that combines these two branches. One of the several modifications is that we do not use \textit{unknown} class predictions for the simplicity of the algorithm. Also, we have an extra non-parametric neck layer, which is inspired by Pang~\etal~\cite{pang2019libra}. They use \textit{balanced semantic features} to enhance the pyramidal neck representations. Different from theirs, our main design purpose is to have a representative feature map itself at a single resolution level. For this reason, our extra neck consists of only the \textit{gather} and \textit{redistribute} steps with no additional parameters. First, at the \textit{gather} step, the input feature pyramid network (FPN)~\cite{lin2017feature} features $\{p^{2}, p^{3}, p^{4}, p^{5}\}$ are resized to the highest resolution \ie, the same size as $p^2$, and element-wise summed over multiple levels, to produce $f$. Then, this representative feature is \textit{redistributed} to the original features by a residual addition.

\paragraph{Fuse at Pixel Level:} The main idea is to leverage video context to improve the per-frame feature by temporal feature fusion. At each time step $t$, the feature extractor is given a target frame $I_{t}$ and one (or more) reference frame(s) $I_{t-\tau}$, then produces FPN features $\{p^{2}, p^{3}, p^{4}, p^{5}\}_{t}$ and $\{p^{2}, p^{3}, p^{4}, p^{5}\}_{t-\tau}$. We sample the reference frame with $\tau \in \{t-5:t+5\}$

We propose an \textit{align-and-attend} pipeline at in between the \textit{gather} and \textit{redistribute} steps. Given the gathered features $f_{t}$ and $f_{t-\tau}$, our \textit{align} module learns flow warping to align the reference feature $f_{t-\tau}$ onto the target feature $f_{t}$. The \textit{align} module receives an initial optical flow $\phi^{init}_{t \rightarrow t-\tau}$ computed by FlowNet2~\cite{ilg2017flownet}, and refine it for more accurate deep feature flow. After concatenating these aligned features, our \textit{attend} module learns spatial-temporal attention to reweight the features and fuse the time dimension into one to get $g_{t}$, which is then redistributed to $\{p^{2}, p^{3}, p^{4}, p^{5}\}_{t}$ which are then fed forward to the downstream instance and semantic branches. 

\paragraph{Track at Object Level:} Here, the goal is to track all object instances in $I_{t}$ with respect to those in $I_{t-\tau}$. Along with the multi-task heads for panoptic segmentation, we add the MaskTrack head~\cite{yang2019video} which is used in a state-of-the-art video instance segmentation method. It learns a $m\times n$ feature affinity matrix $A$ between generated $n$ RoI proposals $\{r_1, r_2...r_n\}_t$ from $I_t$ and $m$ RoI features $\{r_1, r_2...r_m\}_{t-\tau}$ from $I_{t-\tau}$. For each pair $\{r_{i,t}, r_{j,t-\tau}\}$, a Siamese fully-connected layer embeds them into single vectors $\{e_{i,t}, e_{j,t-\tau}\}$, then the cosine similarity is measured by $A_{ij} = cosine(e_{i,t}, e_{j,t-\tau})$.

 MaskTrack is designed for still images and only utilizes appearance features, and does not use any video features during training. To handle this problem, we couple the tracking branch with the temporal fusion module.
Specifically, every RoI features $\{r_1, r_2...r_n\}_t$ are first enhanced by the above temporal fused feature, $g_t$, from multiple frames, and thus become more discriminative before being fed into the tracking branch. Therefore, from a standpoint of the instance tracking, our VPSNet synchronizes it on both pixel-level and object-level. The pixel-level module aligns local feature of the instance to transfer it between the reference and target frames, and the object-level module focuses more on distinguishing the target instance from other reference objects by the similarity function on the temporally augmented RoI features. During training, the tracking head in our VPSNet is the same as ~\cite{yang2019video}. During the inference stage, we add an additional cue from the panoptic head: the IoU of \textit{things} logits. The IoU of instance logits can be viewed as a deformation factor or spatial correlation between frames and our experiments show that it improves the video panoptic quality for \textit{things} classes.

\subsection{Implementation Details}
We follow most of the settings and hyper-parameters of Mask R-CNN and other panoptic segmentation models such as UPSNet~\cite{xiong2019upsnet}. Hereafter, we only explain those which are different. Throughout the experiments, we use ResNet-50 FPN~\cite{he2016deep, lin2017feature} as the feature extractor.

\paragraph{Training: }
We implement our models in PyTorch~\cite{paszke2017automatic} with MMDetection~\cite{chen2019mmdetection} toolbox. We use the distributed training framework with 8 GPUs. Each mini-batch has 1 image per GPU. We use the ground truth box of a reference frame to train the track head. We crop random $800\times1600$ pixels out of  $1024\times2048$ Cityscapes and $1080\times1920$ VIPER images after randomly scaling each frame by 0.8 to 1.25 $\times$. Due to the high resolution of images, we downsample the logits for semantic head and panoptic head to $200\times400$ pixels. Besides the RPN losses, our VPSNet contains 6 task-related loss functions in total: bbox head (classification and bounding-box), mask head, semantic head, panoptic head, and track head. We set all loss weights to 1.0 to make their scales to be roughly on the same order of magnitude.

We set the learning rate and weight decay as 0.005 and 0.0001 for all datasets. For VIPER, we train for 12 epochs and apply lr decay at 8 and 11 epochs. For both Cityscapes and Cityscapes-VPS, we train for 144 epochs and apply lr decay at 96 and 128 epochs. For the pretrained models, we import COCO- or VIPER-pretrained \textit{Base} model parameters and initialize the remaining layers, \eg, Fuse (\textit{align-and-attend}) and Track modules, by Kaiming initialization.

\paragraph{Inference: }
Given a new testing video, our method processes each frame sequentially in an online fashion. At each frame, our VPSNet first generates a set of instance hypotheses. As a mask pruning process, we perform the class-agnostic non-maximum suppression with the box IoU threshold as 0.5 to filter out some redundant boxes. Then the remaining boxes are sorted by the predicted class probabilities and kept if the probability is larger than 0.6. For the first frame of a video sequence, we assign instance ids according to the order of the probability. For all other frames, the remaining boxes after pruning are matched to identified instances from previous frames based on the learned affinity $A$, and are assigned instance id accordingly. After processing all frames, our method produces a sequence of panoptic segmentation, each pixel of which contains a unique category label and instance label throughout the sequence. For both IPQ and VPQ evaluation, we test all available models with single scale testing.

\begin{table}[t]
\centering
\resizebox{0.9\linewidth}{!}{%
\begin{tabular}[b]{ l|ccc|ccc}
        \hline
        Our    & feat. & feat. &  obj.  & PQ & PQ\textsuperscript{Th}  & PQ\textsuperscript{St} \\
        	methods &	align & attend & match & & & \\
        \hline
        Base    &		 & 		&& 52.1 & 47.2 & 56.2 \\
        Align     & \checkmark && & 52.3  & 47.3 & 56.4 \\
        Attend    &  & \checkmark & & 50.7 & 45.8 & 54.8 \\
        Fuse    & \checkmark & \checkmark & & 53.0 & 48.3 & 57.0 \\
        Track     & &  & \checkmark & 53.0 & 47.9 & 57.2 \\
        FuseTrack    & \checkmark & \checkmark &\checkmark & \textbf{55.4} & \textbf{52.2} & \textbf{58.0} \\

        \hline
\end{tabular}
}
\vspace{-2mm}
\caption{Image panoptic segmentation results on VIPER.}
\label{tab:viper_ipq}
\vspace{-2mm}
\end{table}

\begin{table}[t]
\centering
\resizebox{0.9\linewidth}{!}{%
\begin{tabular}[b]{ l|c|ccc}
        \hline
        Method      & Backbone & PQ & PQ\textsuperscript{Th}  & PQ\textsuperscript{St} \\
        \hline
        
        AUNet~\cite{li2019attention}       & ResNet-101         & 59.0 & 54.8 & 62.1 \\
        PanopticFPN~\cite{kirillov2019panoptic2} & ResNet-101   & 58.1 & 52.0 & 62.5 \\       	       		 
        DeeperLab~\cite{yang2019deeperlab}   & Xception-71      & 56.5 & - & - \\
        Seamless~\cite{porzi2019seamless} & ResNet-50           & 59.8 & 54.6 & 63.6 \\
        AdaptIS~\cite{sofiiuk2019adaptis} & ResNet-50           & 59.0 & 55.8 & 61.3 \\
        TASCNet~\cite{li2018learning}       & ResNet-50         & 55.9 & 50.6 & 59.8 \\
        UPSNet~\cite{xiong2019upsnet}       & ResNet-50         & 59.3 & 54.6 & 62.7 \\
        TASCNet+CO~\cite{li2018learning}       & ResNet-50      & 59.2 & 56.0 & 61.5 \\
        UPSNet+CO~\cite{xiong2019upsnet}      & ResNet-50       & 60.5 & 57.0 & 63.0 \\
        \hline
        VPSNet-Base+CO   & ResNet-50 & 60.6    & 57.0 & 63.2 \\
        VPSNet-Fuse+CO     & ResNet-50 & 61.6   & 57.7 & 64.4 \\
        VPSNet-Fuse+VP    & ResNet-50 & \textbf{62.2}   & \textbf{58.0} & \textbf{65.3} \\
        \hline
\end{tabular}
}
\vspace{-2mm}
\caption{Image panoptic segmentation results on Cityscapes \textit{val.} set. `+CO' and `+VP' indicate the model is pretrained on COCO and VIPER, respectively.}
\vspace{-4mm}
\label{tab:city_ipq}
\end{table}

\begin{table*}[]
\centering
\resizebox{\textwidth}{!}{%

\begin{adjustbox}{max width=\textwidth}
\begin{tabular}{l|c|c|c|c| c}
\hline
{VPSNet variants} &\multicolumn{4}{c|}{Temporal window size} 
                & \multirow{2}{*}{VPQ} \\
 \cline{2-5} {on \textbf{VIPER}} & k = 0 & k = 5 & k = 10 & k = 15 &  \\
\hline
Base  & 
			52.1 / 47.2 / 56.2 & 
            29.4 / 0.8 / 53.2 & 
            29.3 / 0.6 / 53.2 &
            29.0 / 0.5 / 52.8 & 	34.9 / 12.3 / 54.1 \\
Fuse  & 
			53.0 / 48.3 / 57.0 & 
            30.0 / 0.8 / 54.4 & 
            29.8 / 0.8 / 54.0 &
            29.6 / 0.6 / 53.8 & 	35.6 / 12.6 / 54.8 \\
Track  & 
			53.0 / 47.9 / 57.2 & 
            47.1 / 39.3 / 53.6 & 
            42.7 / 30.0 / 53.2 &
            40.4 / 25.4 / 52.8 &  45.8 / 35.7 / 54.2 \\

\hline
FuseTrack Cls-Sort & 
			55.4 / 52.2 / 58.0 & 
            30.5 / 0.8 / 55.2 & 
            30.1 / 0.6 / 54.6 &
            29.8 / 0.5 / 54.3 &     36.5 / 13.5 / 55.5 \\
                

FuseTrack IoU-Match  & 
			55.4 / 52.2 / 58.0 & 
            45.0 / 32.8 / 55.2 & 
            40.1 / 22.8 / 54.6 &
            37.9 / 18.2 / 54.3 &    44.6 / 31.5 / 55.5 \\

FuseTrack Disjoined  & 
			55.4 / 52.2 / 58.0 & 
            52.0 / 48.3 / 55.2 & 
            48.6 / 40.4 / 54.6 &
            46.9 / 37.5 / 54.3 &     50.7 / 44.6 / 55.5 \\

FuseTrack (VPSNet)      & 
			\textbf{55.4} / 52.2 / 58.0 & 
            \textbf{53.6} / 51.7 / 55.2 & 
            \textbf{50.1} / 44.7 / 54.6 &
            \textbf{48.4} / 41.4 / 54.3 &   \textbf{51.9} / 47.5 / 55.5 \\
\hline
\end{tabular}
\end{adjustbox}
}
\vspace{-2mm}

\caption{Video panoptic segmentation results on VIPER with our VPSNet variants. Each cell contains VPQ / VPQ\textsuperscript{Th} / VPQ\textsuperscript{St} scores.}
\label{tab:viper_vpq}
\end{table*}

\begin{table*}[]
\centering
\resizebox{\textwidth}{!}{%
\begin{adjustbox}{max width=\textwidth}
\begin{tabular}{l|c|c|c|c| c}
\hline
{VPSNet variants} &\multicolumn{4}{c|}{Temporal window size} 
                & \multirow{2}{*}{VPQ} \\
\cline{2-5} {on \textbf{Cityscapes-VPS \textit{val.}}} & k = 0 & k = 5 & k = 10 & k = 15 &  \\
\hline
Track  & 
			63.1 / 56.4 / 68.0 & 
            56.1 / 44.1 / 64.9 & 
            53.1 / 39.0 / 63.4 &
            51.3 / 35.4 / 62.9 &   
            55.9 / 43.7 / 64.8 \\

FuseTrack (VPSNet) \quad \quad \quad   & 
            \textbf{64.5} / 58.1 / 69.1 & 
            \textbf{57.4} / 45.2 / 66.4 &
            \textbf{54.1} / 39.5 / 64.7 &   
            \textbf{52.2} / 36.0 / 64.0 & 
            \textbf{57.0} / 44.7 / 66.0 \\

\hline
\end{tabular}
\end{adjustbox}
}
\\
\vspace{2mm}
\centering
\resizebox{\textwidth}{!}{%
\begin{adjustbox}{max width=\textwidth}
\begin{tabular}{l|c|c|c|c| c}
\hline
{VPSNet variants} &\multicolumn{4}{c|}{Temporal window size} 
                & \multirow{2}{*}{VPQ} \\
\cline{2-5} {on \textbf{Cityscapes-VPS \textit{test}}} & k = 0 & k = 5 & k = 10 & k = 15 &  \\
\hline
Track  & 
			63.1 / 58.0 / 66.4 & 
            56.8 / 45.7 / 63.9 & 
            53.6 / 40.3 / 62.0 &
            51.5 / 35.9 / 61.5 &     
            56.3 / 45.0 / 63.4 \\

FuseTrack (VPSNet) \quad \quad \quad   & 
			\textbf{64.2} / 59.0 / 67.7 & 
            \textbf{57.9} / 46.5 / 65.1 & 
            \textbf{54.8} / 41.1 / 63.4 &
            \textbf{52.6} / 36.5 / 62.9 &   
            \textbf{57.4} / 45.8 / 64.8 \\

\hline
\end{tabular}
\end{adjustbox}

}
\vspace{-2mm}
\caption{Video panoptic segmentation results on Cityscapes-VPS validation (top) and test (bottom) set with our VPSNet variants. Each cell contains VPQ / VPQ\textsuperscript{Th} / VPQ\textsuperscript{St} scores. Note that while benchmarking our Cityscapes-VPS dataset, we further split our data into 400/50/50 (train/val/test) videos, which result in different performances to those reported in the CVPR 2020 version.}
\vspace{-3mm}
\label{tab:cityvps_vpq}
\end{table*}

\section{Experimental Results}
In this section, we present the experimental results on the two proposed video-level datasets, \textit{VIPER} and \textit{Cityscapes-VPS}, as well as the conventional image-level Cityscapes benchmark. In particular, we mainly investigate the results in two aspects: image-level prediction and cross-frame association, which will be reflected in the IPQ and VPQ, respectively. We demonstrate the contributions of each of the proposed pixel-level Fuse and object-level Track modules in the performance of video panoptic segmentation. Here is the information on the dataset splits used in experiments.

\begin{itemize}[topsep=0.5pt,itemsep=0.5pt]
  \item \textbf{VIPER}: 
 Based on its high quantity and quality of the panoptic video annotation, we mainly experiment with this benchmark. We follow the public train / val split. For evaluation, we choose 10 validation videos from \textit{day} scenario, and use the first 60 frames of each videos: total 600 images. 

  \item \textbf{Cityscapes}:
We use the public train / val split, and evaluate our image-level model on the validation set.

\item \textbf{Cityscapes-VPS}:
The created video panoptic annotations are given with the 500 validation videos of Cityscapes. We further split these videos into 400 training, 50 validation, and 50 test videos. Each video consists of 30 consecutive frames, with every 5 frames paired with the ground truth annotations. For each video, all 30 frames are predicted, and only the 6 frames with the ground truth are evaluated. 

\end{itemize}



\paragraph{Image Panoptic Quality: } 
One thing we can expect from the VPS learning compared to its image-level counterpart is whether it improves per-frame PQ by properly utilizing spatial-temporal features. We evaluate our method with the existing panoptic quality (PQ), recognition quality (RQ), and segmentation quality (SQ). The results are presented in~\tabfref{tab:viper_ipq} and~\tabfref{tab:city_ipq}. 

First, we study the importance of the proposed Fuse and Track modules to our image-level panoptic segmentation performance on the VIPER dataset as shown in~\tabfref{tab:viper_ipq}. We find that both pixel-level and object-level modules have complementary contributions, each improving the baseline by +1\% PQ. Without any of them, the PQ will drop by -3.4\%. The best PQ was achieved when these two modules are used together. 

We also experiment on the Cityscapes benchmark, to provide a comparison with the state-of-the-art panoptic segmentation methods. Our VPSNet with only the Fuse module can be trained in this setting, since it only requires a neighboring reference frame without any extra annotations. In \tabfref{tab:city_ipq}, we find that VPSNet with Fuse module outperforms the state-of-the-art baseline method~\cite{xiong2019upsnet} by +1.0\% PQ, which implies that it effectively exploits spatial-temporal context to improve per-frame panoptic quality. The pretraining on the VIPER dataset shows its complementary effectiveness to either COCO or Cityscapes dataset by boosting the score by +1.6\% PQ from our baseline, achieving 62.2\% PQ. We also converted our results into \textit{semantic} segmentation format, and achieved 79.0\% mIoU.

 \begin{figure*}[t]
\begin{center}
\begin{tabular}{@{}c@{}}
\includegraphics[width=\linewidth]{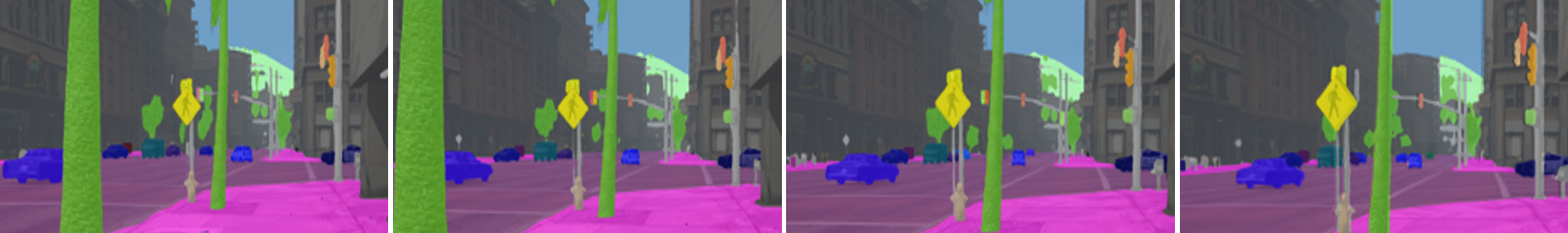} \\
\includegraphics[width=\linewidth]{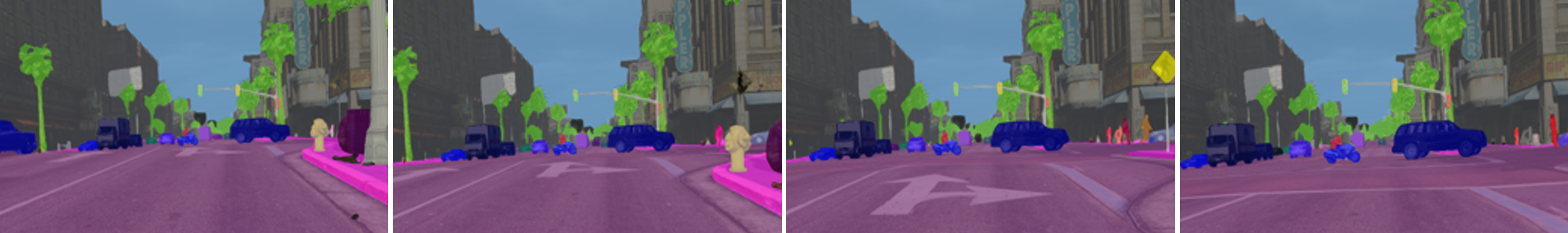} \\
\includegraphics[width=\linewidth]{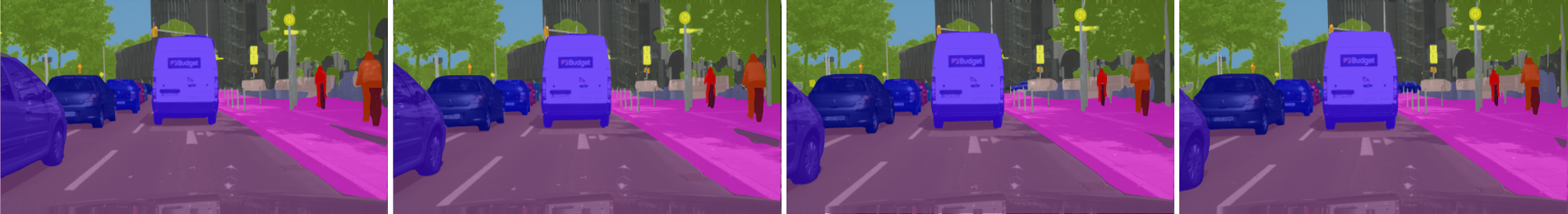} \\
\includegraphics[width=\linewidth]{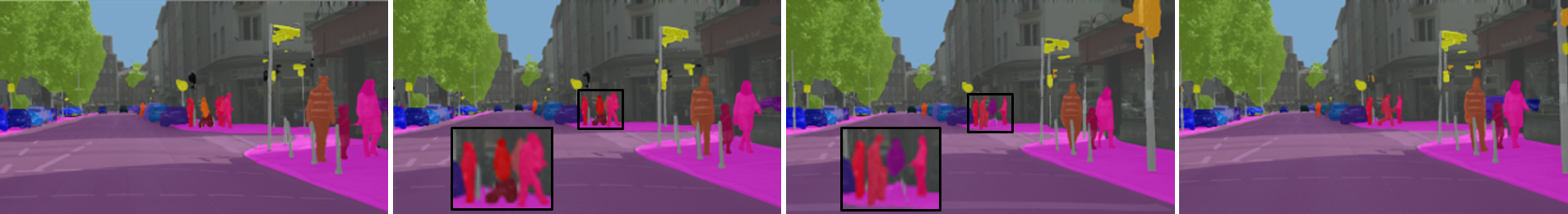} 
\end{tabular}
\end{center}
\vspace{-5mm}
\caption{\textbf{Sample results of VPSNet on VIPER and Cityscapes-VPS.} Each row has four sampled frames from a video sequence of VIPER (top two rows) and Cityscapes-VPS (bottom two rows). The last row includes failure cases when the crowded objects are crossing each other. Objects with the same predicted identity have the same color.}
\label{fig:qual}
\end{figure*}

\paragraph{Video Panoptic Quality: }
We evaluate the spatial-temporal consistency between the predicted and ground truth panoptic video segmentation. The quantitative results are shown in \tabfref{tab:viper_vpq} and \tabfref{tab:cityvps_vpq}. Different from image panoptic segmentation, our new task requires extra consistency in instance ids over time, which makes the problem much more challenging for \textit{things} than stuff classes. Not surprisingly, the mean video panoptic quality of things classes (VPQ\textsuperscript{Th}) is generally lower than that of stuff classes (VPQ\textsuperscript{St}).

Since there is no prior work directly applicable to our new task, we present several baseline VPS methods to provide a reference level. Specifically, we enumerate over different methods by replacing only the tracking branch of our VPSNet. The alternative tracking methods are object sorting by classification logit values (Cls-Sort), and flow-guided object matching by mask IoU (IoU-Match). First, Cls-Sort relies on semantic consistency of the same object between frames. However, it fails to track objects possibly because there are a number of instances of the same class in a frame, \eg, car, person, thus making the class logit information not enough for differentiating these instances. On the other hand, IoU-Match is a simple yet strong candidate method for our task by leveraging spatial correlation to determine the instance labels, improving the image-level baseline by +9.7\% VPQ. 

Our model with Track module further improves this by +1.2\% VPQ, by using the learned RoI feature matching algorithm together with the semantic consistency and spatial correlation cues. Our full model with both Fuse and Track modules achieves the best performance by a great improvement of  +17.0\% over the image-level base model, and +6.1\% VPQ over the variant with only-Track module. To show the contribution of the fused feature solely on the object matching performance, we experiment with a VPSNet variant where the fused feature is fed to all task branches except for the tracking branch (Disjoined). The result implies that the Fuse and Track modules share information, and synergize each other to learn more discriminative features for both segmentation and tracking. We observed the consistent tendency with our Cityscapes-VPS dataset, where our full VPSNet (FuseTrack) achieves +1.1\% VPQ higher than the Track variant. \figureref{fig:qual} shows the qualitative results of our VPSNet on VIPER and Cityscapes-VPS.


\paragraph{Discussion: }
We find several challenges still remaining for our new task. First, even the state-of-the-art video instance tracking algorithm~\cite{yang2019video} and our VPSNet suffer a considerable performance drop as the temporal length increases. In the context of video, possible improvements are expected to made on handling a large number of instances and resolving overlaps between these objects, \eg, \figureref{fig:qual}-(4th row), by better modeling the temporal information~\cite{oh2019video,zhu2017deep}. Second, our task is still challenging for \textit{stuff} classes as well considering the fact that the window size of 15 frames represents only $0.5 \sim 1$ second in a video. The mutual exclusiveness between things and stuff class pixels could be further exploited to encourage both semantic segmentation and instance segmentation to regularize each other.

Another important future direction is to improve the efficiency of an algorithm as in several video segmentation approaches~\cite{li2018low,shelhamer2016clockwork} by sampling keyframes and propagate information in between to produce temporally dense panoptic segmentation results.

\section{Conclusion}
We present a new task named video panoptic segmentation with two types of associated datasets. The first is to adapt the synthetic VIPER dataset into our VPS format, which can provide maximal quantity and quality of panoptic annotations. The second is to create a new video panoptic segmentation benchmark, \textit{Cityscapes-VPS} which extends the popular image-level Cityscapes dataset. We also propose a new method, VPSNet, by combining the temporal feature fusion module and object tracking branch with a single-frame panoptic segmentation network. Last but not least, we suggest a video panoptic quality measure for evaluation to provide early explorations towards this task. We hope the new task and new algorithm will drive the research directions to step forward towards video understanding in the real-world.

\paragraph{Acknowledgements}
This work was in part supported by the Institute for Information \& Communications Technology Promotion (2017-0-01772) grant funded by the Korea government. Dahun Kim was partially supported by Global Ph.D. Fellowship Program through the National Research Foundation of Korea (NRF) funded by the Ministry of Education (NRF-2018H1A2A1062075).

{\small
\bibliographystyle{ieee_fullname}
\bibliography{egpaper_arxiv.bbl}
}
\clearpage
\begin{figure*}[t]
\begin{center}
\begin{tabular}{@{}c@{}}

\includegraphics[width=\linewidth]{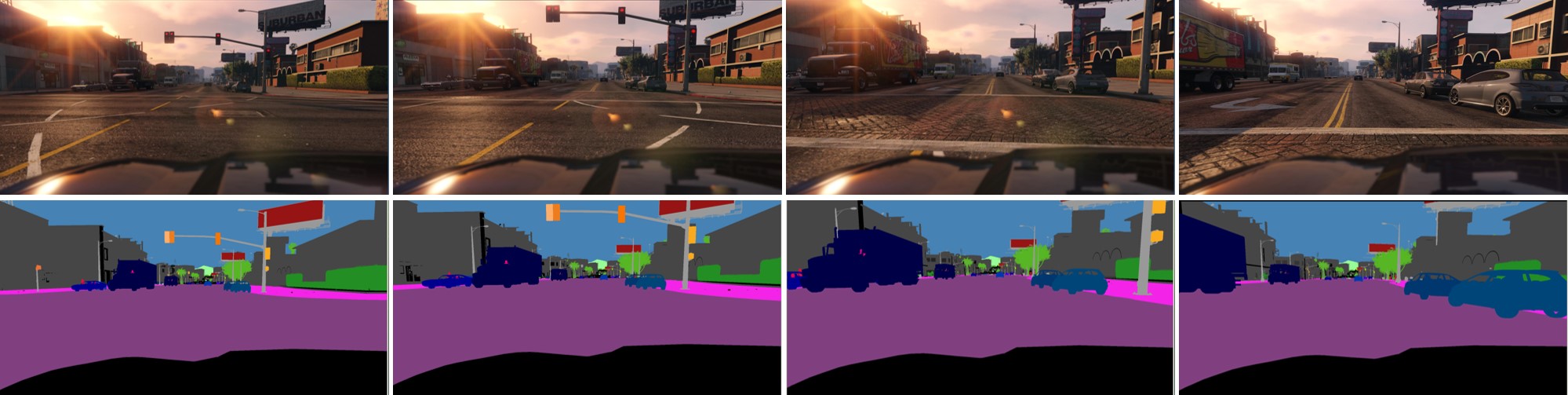} \\ \\
\includegraphics[width=\linewidth]{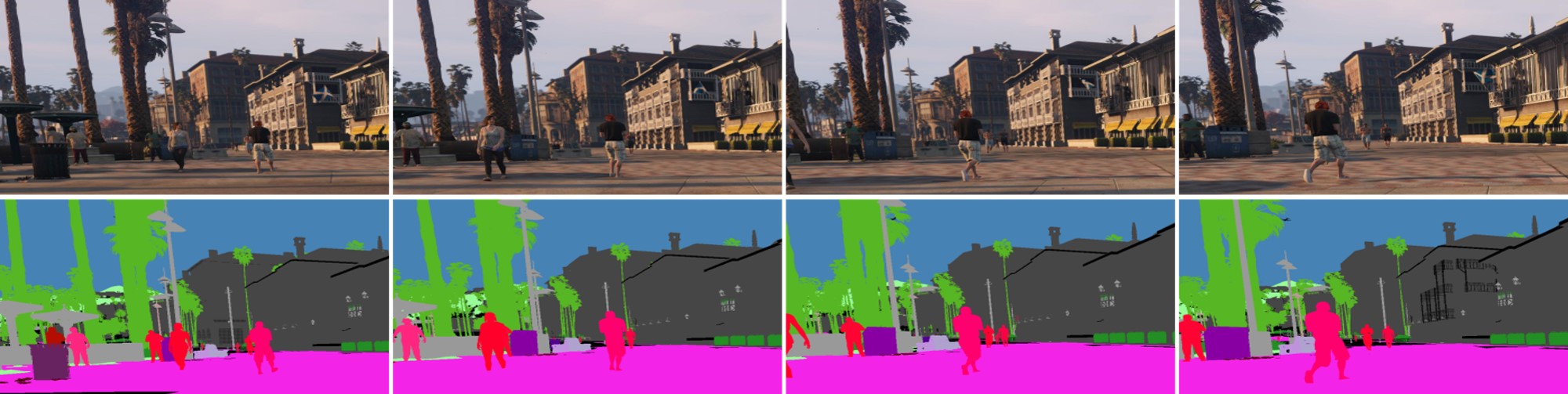} \\ \\
\includegraphics[width=\linewidth]{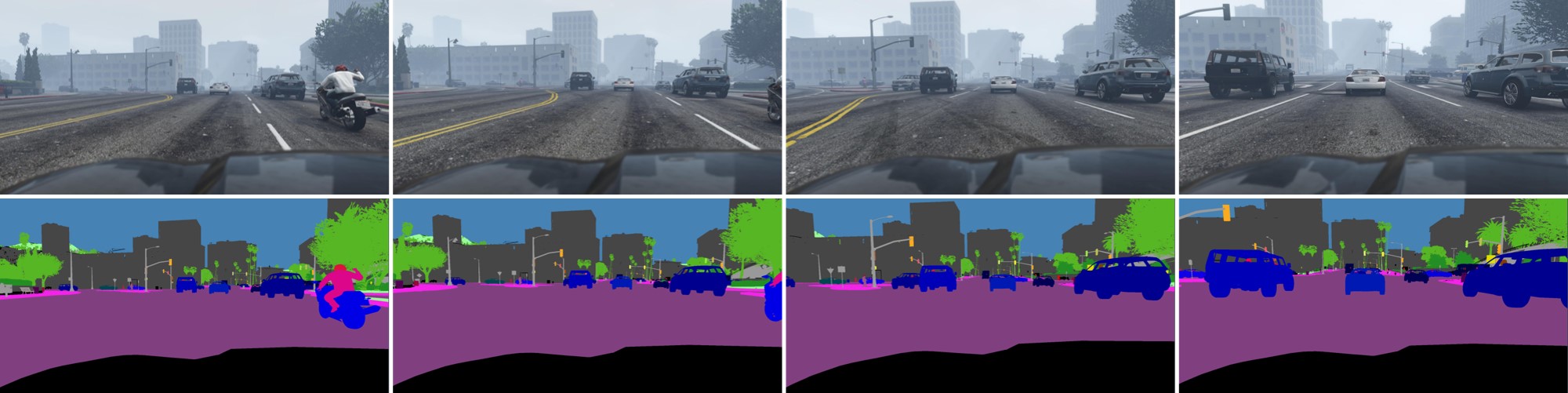} \\ \\
\includegraphics[width=\linewidth]{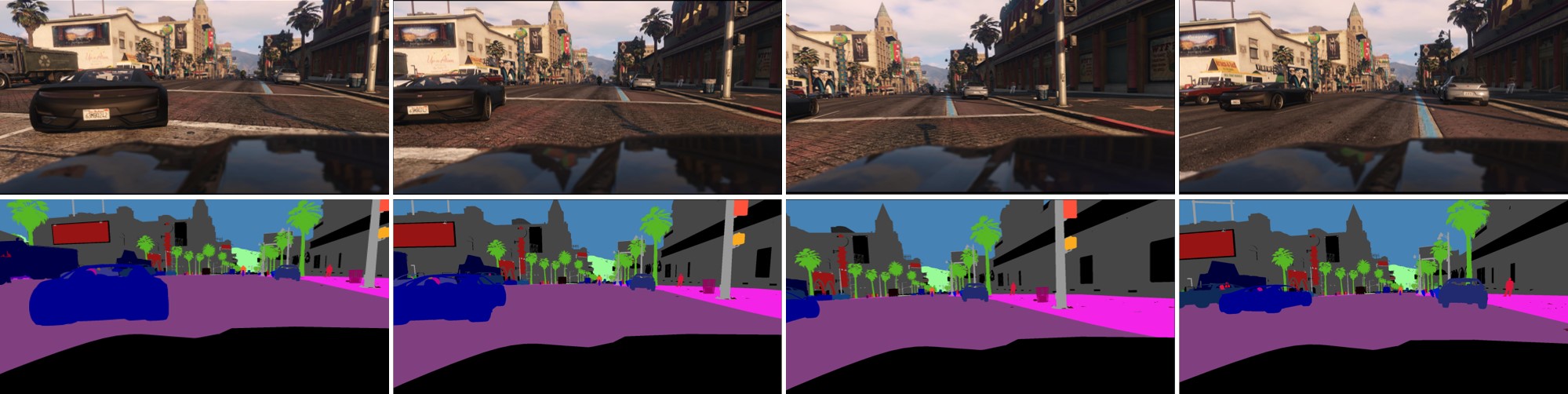} 

\end{tabular}
\end{center}
\caption{\textbf{Sample video sequences of our reformatted VIPER annotations}. Objects of a same semantic class have similar color, where the colors of each instances are randomly deviated. Objects with the same identity have the same color across frames.}
\label{fig:viper}
\end{figure*}

\begin{figure*}[t]
\begin{center}
\begin{tabular}{@{}c@{}}
\includegraphics[width=\linewidth]{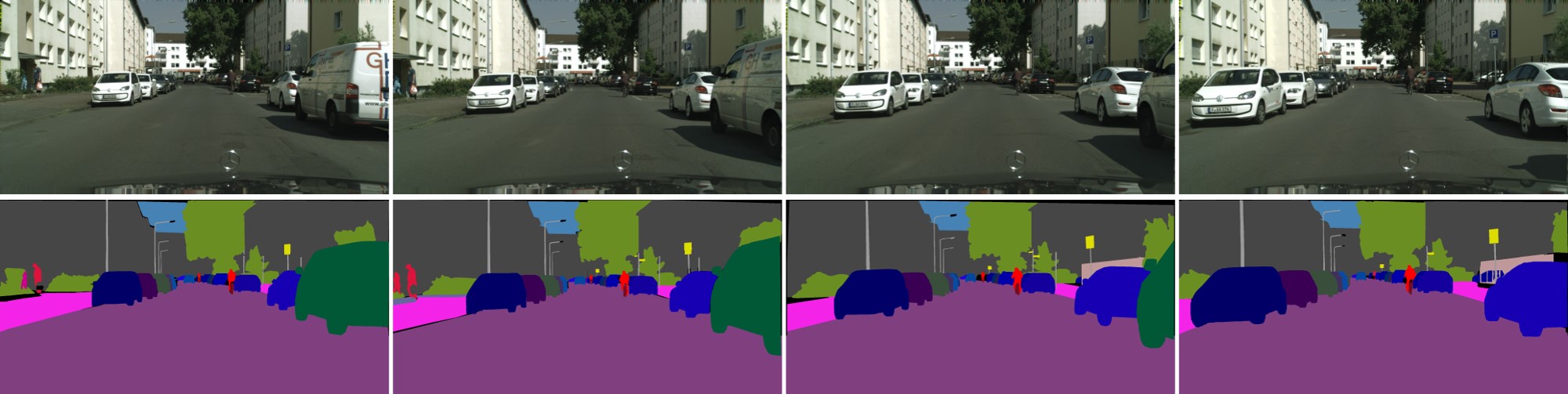} \\ \\
\includegraphics[width=\linewidth]{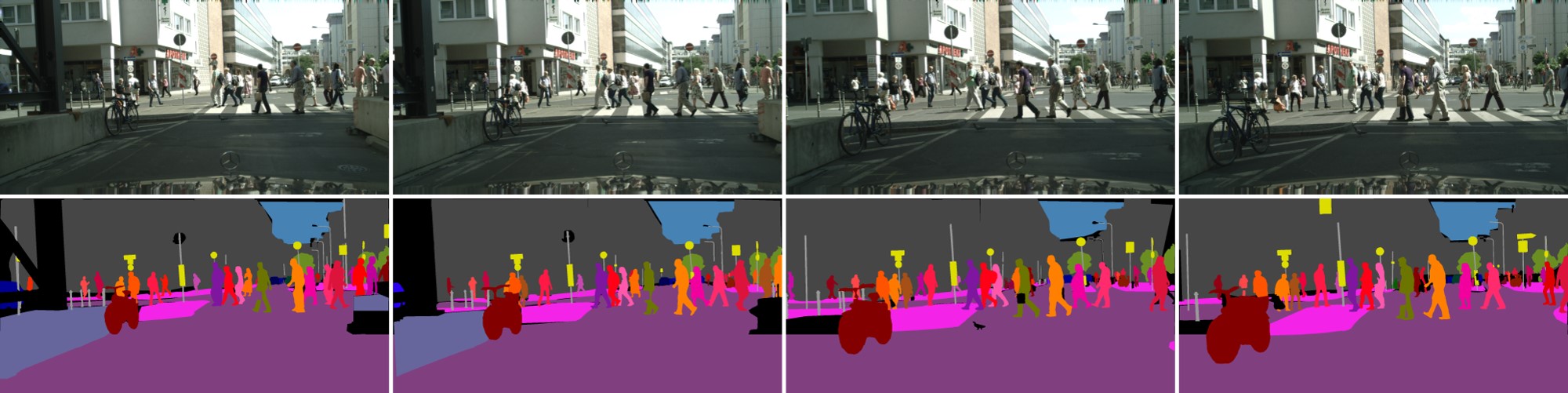} \\ \\
\includegraphics[width=\linewidth]{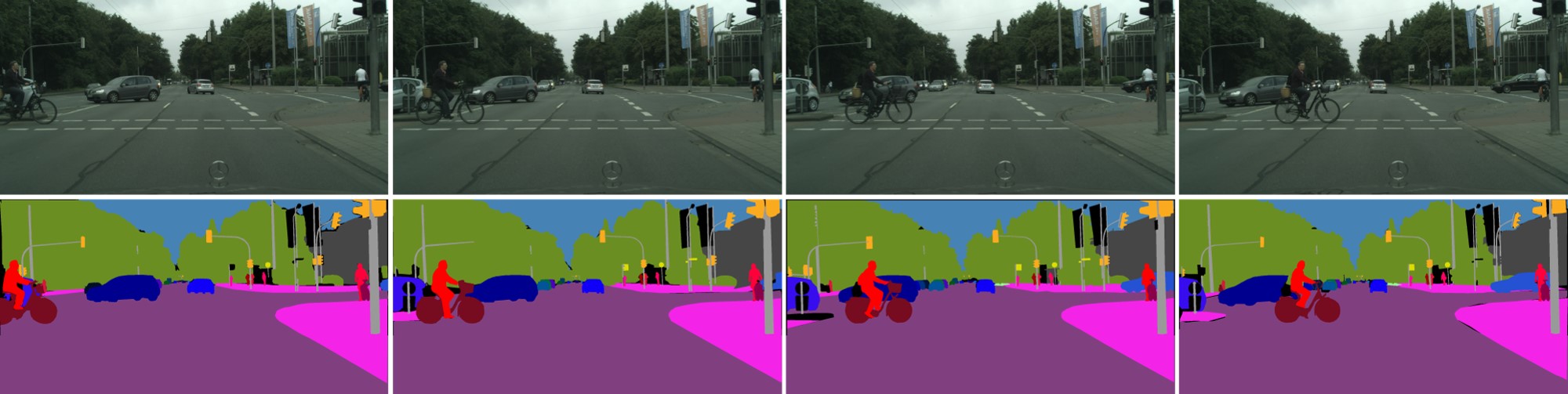} \\ \\
\includegraphics[width=\linewidth]{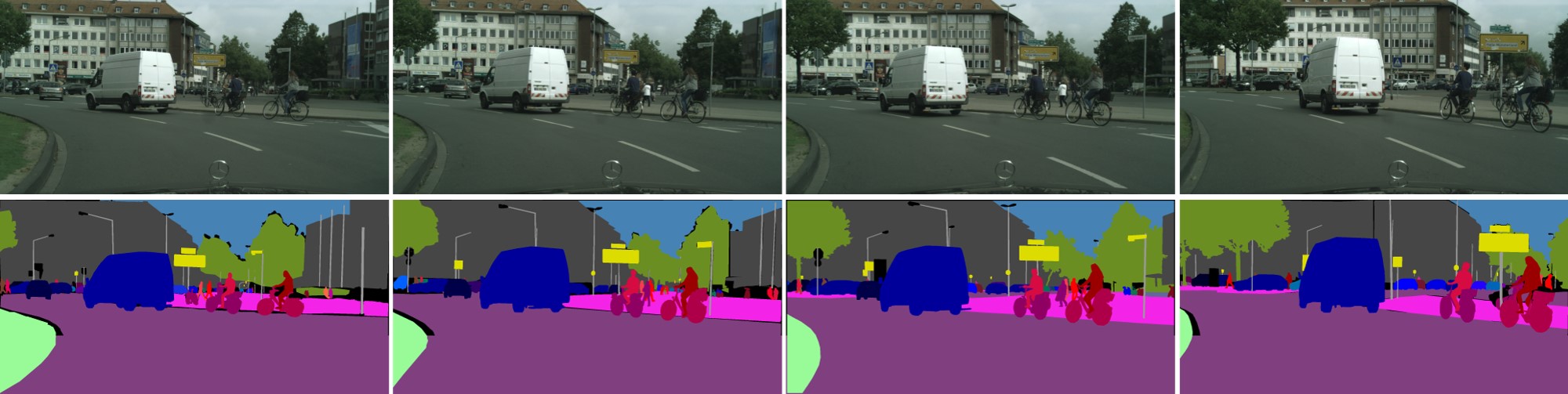} 
\end{tabular}
\end{center}
\caption{\textbf{Sample video sequences of our created Cityscapes-VPS annotations}. Objects of a same semantic class have similar color, where the colors of each instances are randomly deviated. Objects with the same identity have the same color across frames.}
\label{fig:city}
\end{figure*}

\end{document}